\newif\ifshowfigs
\showfigstrue

\providecommand{\doi}[1]{https://doi.org/#1}

\documentclass[10pt,letterpaper]{article}
\usepackage{hyperref}
\usepackage[top=0.85in,left=2.75in,footskip=0.75in]{geometry}

\ifshowfigs
    \usepackage[pdftex]{graphicx}
    \DeclareGraphicsExtensions{.eps,.tiff,.png}
    \usepackage{epstopdf}
    \AppendGraphicsExtensions{.tiff}
\fi

\usepackage{amsmath,amssymb}

\usepackage{changepage}

\usepackage[utf8x]{inputenc}

\usepackage{textcomp,marvosym}

\usepackage{cite}

\expandafter\def\expandafter\UrlBreaks\expandafter{\UrlBreaks
  \do\a\do\b\do\c\do\d\do\e\do\f\do\g\do\h\do\i\do\j%
  \do\k\do\l\do\m\do\n\do\o\do\p\do\q\do\r\do\s\do\t%
  \do\u\do\v\do\w\do\x\do\y\do\z\do\A\do\B\do\C\do\D%
  \do\E\do\F\do\G\do\H\do\I\do\J\do\K\do\L\do\M\do\N%
  \do\O\do\P\do\Q\do\R\do\S\do\T\do\U\do\V\do\W\do\X%
  \do\Y\do\Z}

\usepackage[right]{lineno}

\usepackage{microtype}
\DisableLigatures[f]{encoding = *, family = * }

\usepackage[table]{xcolor}

\usepackage{array}

\newcolumntype{+}{!{\vrule width 2pt}}

\newlength\savedwidth

\raggedright
\usepackage[aboveskip=1pt,labelfont=bf,labelsep=period,justification=raggedright,singlelinecheck=off]{caption}

\makeatletter
\renewcommand{\@biblabel}[1]{\quad#1.}
\makeatother

\usepackage{lastpage,fancyhdr,graphicx}
\usepackage{epstopdf}
\fancyheadoffset[L]{2.25in}
\fancyfootoffset[L]{2.25in}
\begin{document}
\vspace*{0.2in}

\begin{flushleft}
{\Large
\textbf\newline{A deep learning approach for lower back-pain risk prediction during manual lifting} 
}
\newline
\\
Kristian Snyder\textsuperscript{1*},
Brennan Thomas\textsuperscript{1},
Ming-Lun Lu\textsuperscript{2\Yinyang},
Rashmi Jha\textsuperscript{1\Yinyang},
Menekse S. Barim\textsuperscript{2\ddag},
Marie Hayden\textsuperscript{2\ddag},
Dwight Werren\textsuperscript{2\ddag}
\\
\bigskip
\textbf{1} Department of Electrical Engineering and Computing Systems, University of Cincinnati, Cincinnati, Ohio, United States of America
\\
\textbf{2} National Institute for Occupational Safety and Health, Cincinnati, Ohio, United States of America
\\
\bigskip

%
%
\Yinyang These authors contributed equally to this work.

\ddag These authors also contributed equally to this work.

\textcurrency Current Address: Department of Engineering and Computer Science, University of Cincinnati, Cincinnati, Ohio, United States of America 

* snyderks@mail.uc.edu

\end{flushleft}
\section*{Abstract}
Occupationally-induced back pain is a leading cause of reduced productivity in industry. Detecting when a worker is lifting incorrectly and at increased risk of back injury presents significant possible benefits. These include increased quality of life for the worker due to lower rates of back injury and fewer workers' compensation claims and missed time for the employer.
However, recognizing lifting risk provides a challenge due to typically small datasets and subtle underlying features in accelerometer and gyroscope data. 
A novel method to classify a lifting dataset using a 2D convolutional neural network (CNN) and no manual feature extraction is proposed in this paper; the dataset consisted of 10 subjects lifting at various relative distances from the body with 720 total trials. 
The proposed deep CNN displayed greater accuracy (90.6\%) compared to an alternative CNN and multilayer perceptron (MLP). 
A deep CNN could be adapted to classify many other activities that traditionally pose greater challenges in industrial environments due to their size and complexity.

\section*{Introduction}
Back pain, especially when occupationally-induced, is an extremely common ailment. 23.2\% of the world’s population is estimated to be affected in any given month \cite{Hoy2012}. It is considered the leading cause of job-related disability and missed work days, resulting in massive losses of productivity \cite{bernhard1997critical, national2001musculoskeletal}. Back pain is, in the United States, the largest contributor to total workers’ compensation costs as of 2015, representing over 20\% of costs and 13.7 billion USD annually \cite{LibertyMutualResearchInstituteforSafety2018}.

The revised National Institute for Occupational Safety and Health (NIOSH) lifting equation \cite{Waters1994} (RNLE) is currently considered the leading method of measuring back pain risk involved in both single and repeated lifting of objects in the workplace \cite{waters2011, dempsey2005}. Given the mass, relative source distance, relative destination distance (to the person) of an object, frequency of the lifting tasks, and the work-rest pattern, the RNLE determines the relative level of risk to the lifter. Taking measurements for using the RNLE in the field presents a challenge because the analyst needs to interrupt work activity for measuring several lifting variables to recognize the characteristics of lifting tasks. In a world of activity recognition and wearable sensors becoming prominent \cite{Music2013, Liu2016, Bao2004, Preece2009}, there is a distinct lack of research into automatically detecting lift risk, especially as it relates to back pain.

Real-time recognition of unsafe lifting would provide immediate feedback to the users, something currently impractical with traditional methods such as a specialist constantly monitoring the workers. Therefore, the ability to, in real-time, detect unsafe lifting behavior in an industrial setting would provide significant benefits. Closing the feedback loop allows for the wearer to be quickly alerted to any risk and help prevent further stress. A response time in seconds presents a significant improvement over typical pain feedback, which does not always present itself quickly and may be alongside debilitating injury \cite{Mayer2006}. Additionally, an automated approach is far more scalable and can be inexpensively rolled out to an entire set of workers for minimal cost compared to treatment and loss of productivity due to back injury.

Classifying a person lifting objects of various distance from themselves presents an inherent challenge because even to an observer, the different movements are quite similar. Additionally, sourcing data for specialized activities has its own challenges, given a dearth of existing datasets and the increased expense of independently collecting data. Consequently, datasets are typically small; the NIOSH lifting dataset consists of 720 total trials \cite{Barim2019}. The challenge, then, is to develop a versatile model that can distinguish between very similar activities and operates on datasets possibly magnitudes smaller than for similar problems \cite{Weiss2016}.

Significant advancements in deep learning approaches have been achieved in recent years, broadly categorized into video-based and sensor-based classification \cite{Cook2013}. Video-based approaches have focused mostly on surveillance and gait recognition, although there is recent research that has successfully labeled live video data with a high degree of accuracy without assistance from manual body part labeling or tracking devices \cite{Cook2013, Jalal2017}. While impressive, use of video classification in many workplaces requires a wide array of surveillance cameras placed to view all workers, which can be expensive and require complicated installation. In other areas, such as most construction and industrial sites, video is near-impossible due to the frequently temporary and constantly fluctuating workplaces. Sensors placed on the body do not require the person to be in a specific place and are relatively inexpensive and less privacy-intrusive than constant video capture  \cite{Cook2013}, making them more practical for this use case.

Leading sensor-based approaches typically classify more distinct activities, such as standing, walking, and running with relative success \cite{Kuspa2013, Kwapisz2011, Weiss2016}. However, classification of visually similar activities (such as walking up and down stairs) displays notably lower performance \cite{Kuspa2013}. Additionally, other attempts to classify accelerometer data focused on $n>500$ samples for each class and commonly in excess of thousands of samples \cite{Bao2004, Kuspa2013, Weiss2016, Maurer2006, Hammerla2016}. The existence of other datasets with multiple accelerometers suggests a place for a model that can work on very detailed activities generating much more data with dedicated sensors than a smartwatch or smartphone \cite{Bao2004}. 

In this paper, a model utilized to classify the above lifting activities is presented as a generalizable solution to the problem of classifying small datasets of very similar activity classes. The model uses 2-dimensional convolutional layers along with average pooling to classify the activities. Accelerometer and gyroscope data is first preprocessed with a Butterworth filter and manipulated into a 3D matrix resembling an image before being trained on. The structure of the model is examined by comparing it to variations in pooling, regularization, and complexity to display the theoretical underpinnings, increasing its adaptability to other problems. The network's various hyperparameters are also examined to justify their specific values. Finally, accuracy and other statistics of the proposed model are compared to traditional activity recognition models and their published results.

Expanding the list of activities to a wider subset of possible movement unlocks a multitude of possible benefits. In collaboration with the National Institute for Occupational Safety and Health (NIOSH), a deep convolutional neural network (CNN) model to classify relative risk level of lifting objects is developed. Such a model would assist in preventing serious, chronic back injury in workers and significantly improve their quality of life.

\section*{Materials and methods}
\subsection*{Data collection}

\begin{figure}[!ht]
    \ifshowfigs
    \includegraphics{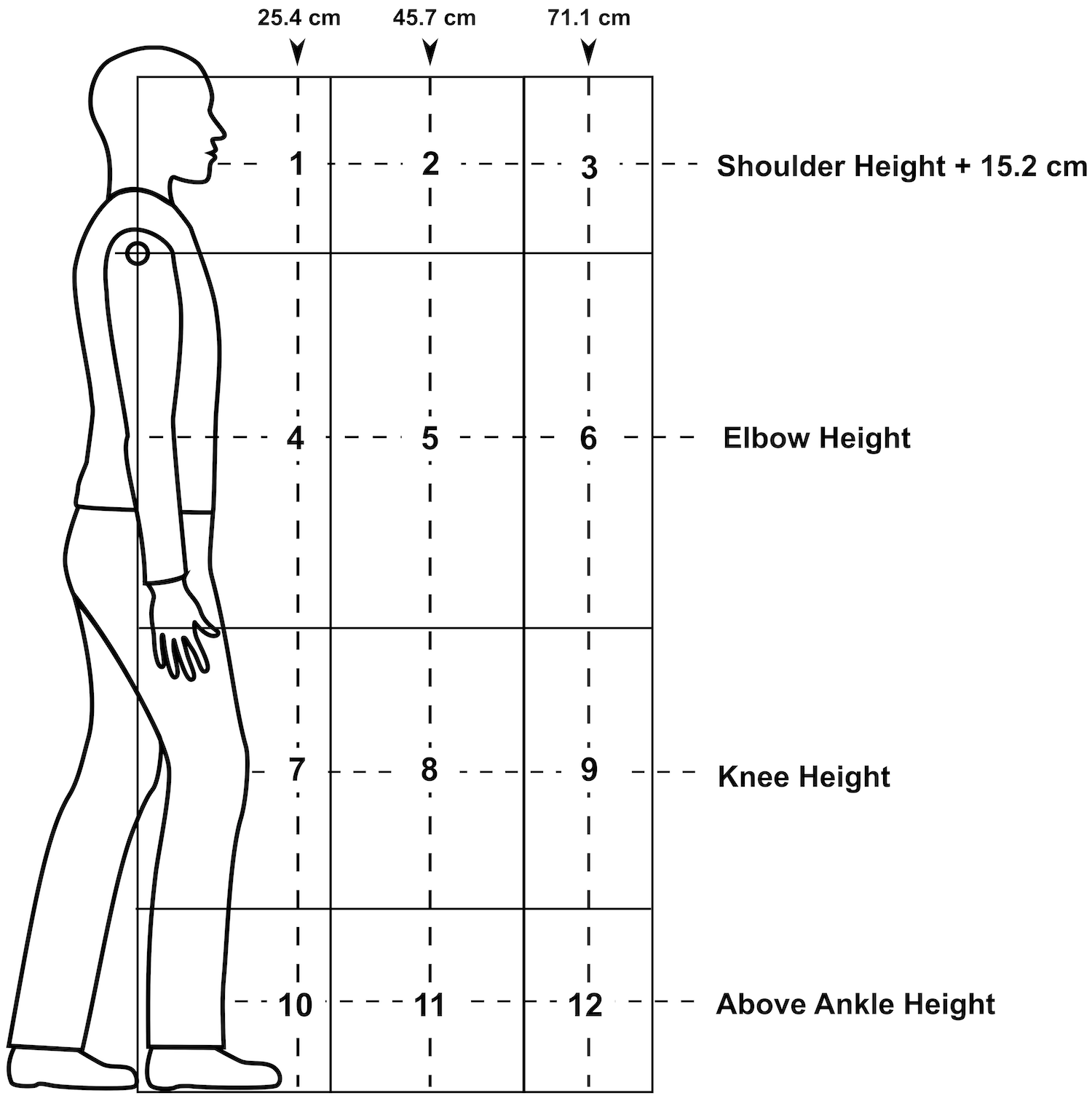}
    \fi
    \caption{{\bf ACGIH lifting zone system depicting the relative areas collected for analysis.} (Source: NIOSH)}
    \label{fig:TLVzones}
\end{figure}

For purposes of analysis, data utilized for model development was sourced from a previous study by researchers at NIOSH to examine body motion for two-handed lifting tasks similar to those in the workplace \cite{Barim2019}. Lifts were performed using the American Conference of Governmental Industrial Hygienists (ACGIH) Threshold Limit Values (TLV) for lifting, which defines 12 zones relative to the body in the sagittal plane shown in Fig~\ref{fig:TLVzones}.
Save for Zones 1-3, 4, 7, and 10, all lifts began in the midpoint of these zones. In the aforementioned exceptions, initial points in the zone were altered to provide realistic motion for the subjects' ranges of motion. The object lifted consisted of a $36 \text{cm} \times 12$ cm wire grid weighing 0.45 kg with two handles, simulating a crate or box. To prevent injury, the weight was kept small. The order of lifting from each zone was randomized to prevent bias from ordering the lifts. A total of 720 trials with 6 trials for each subject in all zones were available as input data.


\begin{figure}[!ht]
    \ifshowfigs
    \centering
    \includegraphics[height=6in]{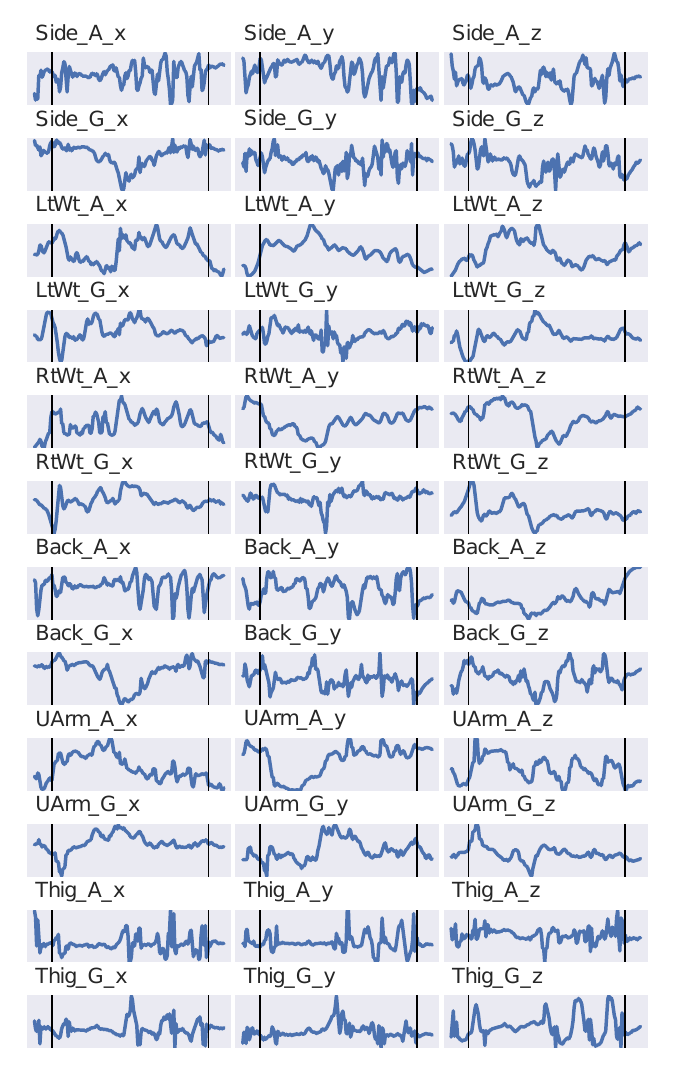}
    \fi
    \caption{{\bf Plots of the accelerometer and gyroscope sensors for Subject 1's first lift in Zone 1 (high risk) from \cite{Barim2019}.} Not all data collected is shown; the two vertical black lines show the beginning and end of the actual lift.}
    \label{fig:example_plots}
\end{figure}

Five male and five female subjects (mean and SD: $170 \pm 7.4$ cm for height and $85.7 \pm 20.2$ kg for weight) participated in the lifting process. All participants had six inertial measurement unit (IMU) sensors (Kinetic Inc.) attached to their bodies on the upper back (T12), each wrist, the dominant upper arm, waist, and the dominant thigh during all lifting. Each sensor consisted of a tri-axial gyroscope and accelerometer sampling at a rate of 25 Hz. All sensor data was calibrated and synchronized prior to data collection \cite{Barim2019}. An example of this data is shown in Fig~\ref{fig:example_plots}.

\subsection*{Feature extraction}

In training the model, 540 of the 720 total trials were utilized for training while the remaining 25\% (180 trials) formed the test set. Training and testing subsets were sampled randomly without replacement. Features were the preprocessed lift, with zero-padding applied to lifts that did not reach the full time. All training utilized all of the 6 sensor areas, each containing an accelerometer and gyroscope, resulting in 18 total measurements for each point in time. The twelve ACGIH lifting zones were mapped to three risk levels: low, medium, and high-risk. The mapping is based on a slight modification to the Los Alamos National Laboratory recommendations to simplify the ACGIH zones; the zones mapped are listed in Table~\ref{table1}.

\begin{table}[!ht]
\centering
\caption{
{\bf Mapping of ACGIH lifting zones to relative risk levels.}}
\label{table1}
\begin{tabular}{ll}
\textbf{Risk Level } & \textbf{Zones }      \\ 
\hline
Low                  & 4, 5                 \\ 
\hline
Medium               & 6, 7, 8, 9           \\ 
\hline
High                 & 1, 2, 3, 10, 11, 12  \\
\hline
\end{tabular}
\end{table}

Finally, in each case the data was scaled with normalization to the bounds [-1, 1] based on the training data, with the same scaling applied to the testing data. This is in direct interest of increasing performance of the neural network, which trains best on data normalized to these bounds.
This resulted in 720 total 27,000-dimension class-labelled vectors for training and testing.

A maximum time window of 30 seconds was selected to train the model. Lifts that did not reach this length of time were zero-padded to reach the full time. Most trials did not reach this time period; the majority ended between 10 and 15 seconds. The long time slice length was selected to investigate dependence on alignment of the starting and ending times, ideally mitigating or eliminating any significant dependence. To reduce overall noise and drift, a Butterworth filter with order 2, lower bound of 2 Hz and upper bound of 12 Hz was applied to each dimension (X, Y, Z) of the gyroscope and accelerometer in each sensor, resulting in 36 total measurements for each point in time. A Butterworth filter is a signal processing filter that has maximally flat frequency response in the passband, preserving the original signal better than other filters \cite{Erer2007}. Each dimension of each of the 12 sensors (an acceleromter and gyroscope each on the side, left wrist, right wrist, back, upper arm, and thigh) has 750 data points in each trial, resulting in 27,000 total features as shown in Eq~\ref{eq:features}.

\begin{equation}
\label{eq:features}
\centering
\begin{split}
& \quad \text{6 IMU devices} \\
\times & \quad \text{2 sensors per IMU (gyro and accelerometer)} \\
\times & \quad \text{3 dimensions each} \\ 
\times & \quad \text{30 sec max per trial} \\
\times & \quad \frac{\text{25 frames}}{\text{sec}} \\
= & \quad \text{27,000 features}
\end{split}
\end{equation}

Parameters and the filter itself were chosen based on previous research \cite{kwon2011validation}, \cite{Mayagoitia2002}, \cite{Music2013}.

\begin{figure}[!hbtp]
\vspace{-27pt}
\begin{adjustwidth}{-2.25in}{0in}
    \centering
    \ifshowfigs
    \includegraphics[height=\textheight]{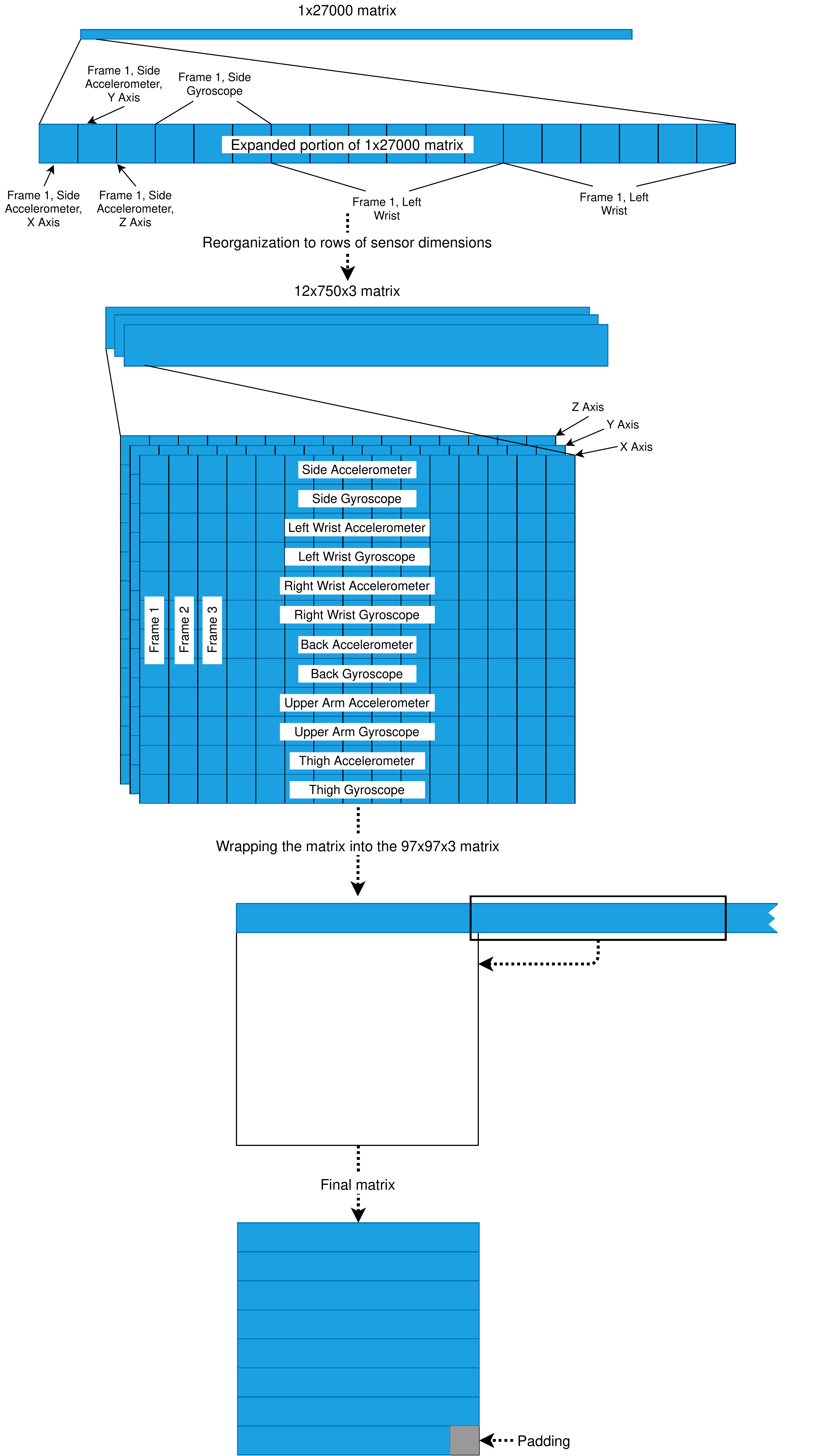}
    \fi
    \caption{{\bf Process of ingesting data into the model for training.} All figures not to scale. The resultant matrix could theoretically be any size; a square was selected for highest compatibility with existing CNN research.}
    \label{fig:ingestionflow}
\end{adjustwidth}
\end{figure}

\begin{figure}[!ht]
    \ifshowfigs
    \includegraphics[width=\textwidth]{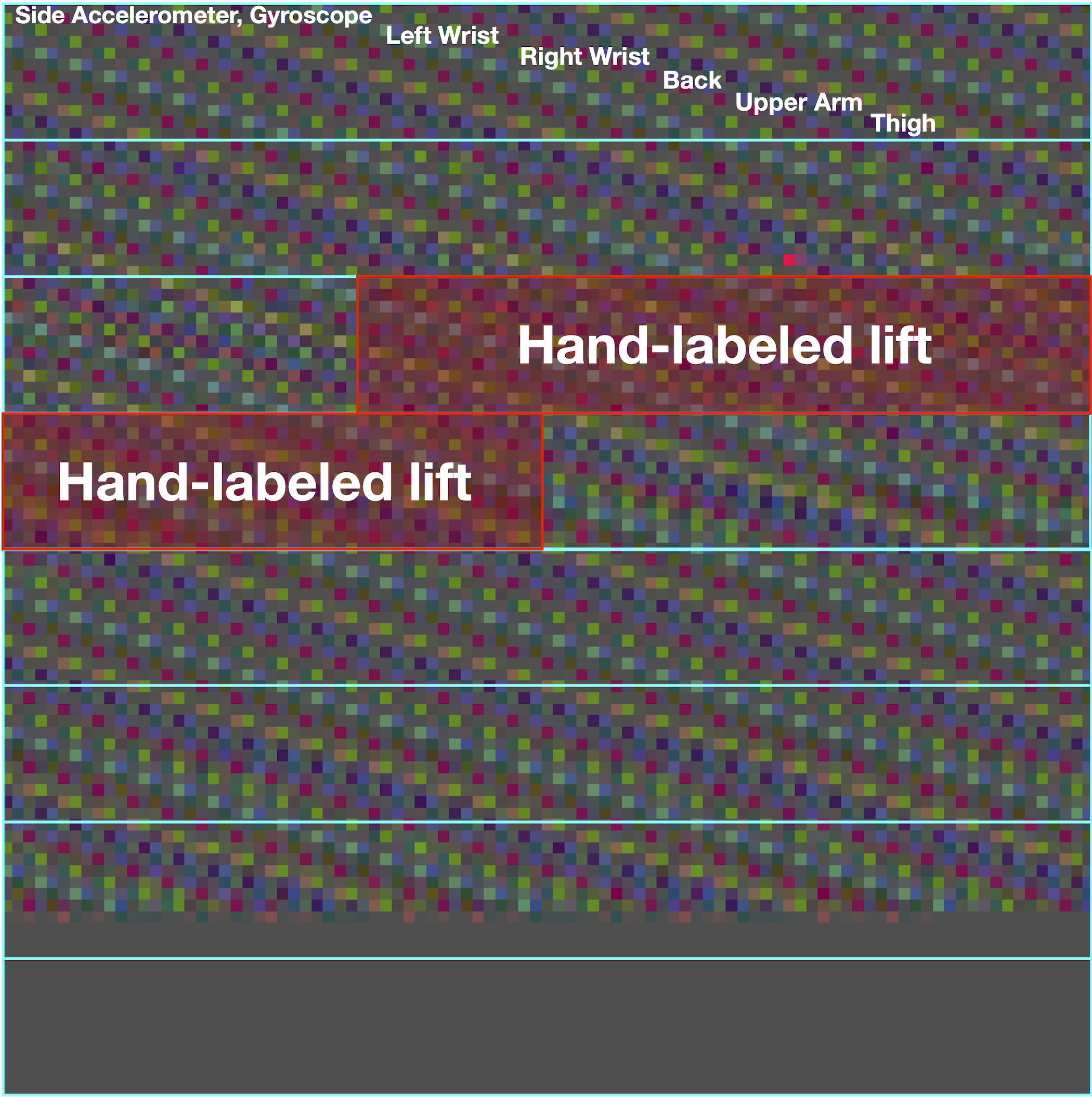}
    \fi
    \caption{{\bf Example of an input image to the network.} The image shown has had a Butterworth bandpass filter of order 2 and bounds 2 and 12 Hz applied to it in addition to a standardizing scaler. The grey block at the bottom represents padding to the model that makes all inputs the same size.}
    \label{fig:squaredimage}
\end{figure}

To prepare the data for ingestion to the model, each 27,000-feature vector was reshaped to form a $95 \times 95 \times 3$ matrix. Specific reshaping is performed by stacking each sensor to form a $36 \times 750 \times 3$ matrix, placing each point of time into successive columns. The matrix is then line-wrapped to form the final image. See Fig~\ref{fig:ingestionflow} for a visualization of the process. Fig~\ref{fig:squaredimage} displays an example of the final image after standardization.

This method produces a data format friendly to CNN models while preserving time locality of the data as much as possible. Convolutions, therefore, will be made more often between features occurring at similar times to help correct for an imprecise window mislabeling the start and end of the lift. Finally, the data is standardized for each sensor by scaling the data to a mean of zero and standard deviation of one.

\subsection*{Overall statistics}

During testing and training of the proposed model, several class-specific statistics were collected to help measure its performance. In addition to the statistics below, two overall statistics were also collected: R\textsubscript{K} (primarily utilized in hyperparameter tuning, see 3.1. Hyperparameter tuning) and overall accuracy.

Precision (shown in Eq~\ref{eq:precision}) is defined as the proportion of instances belonging to the class (true positive or TP) over all instances (both TP and false positive or FP) classified as that class.

\begin{eqnarray}
\label{eq:precision}
	Precision = \frac{TP}{TP + FP}
\end{eqnarray}

Recall (shown in Eq~\ref{eq:recall}) is defined as the proportion of a elements in a particular class classified as that class over all elements belonging to that particular class (including TP and false negative or FN).

\begin{eqnarray}
\label{eq:recall}
	Recall = \frac{TP}{TP + FN}
\end{eqnarray}

F-measure (shown in Eq~\ref{eq:f-measure}) is the harmonic mean of precision and recall, used as an alternative to raw accuracy.

\begin{eqnarray}
\label{eq:f-measure}
	\textit{F-measure} = \frac{2 \times Recall \times Precision}{Recall + Precision}
\end{eqnarray}

\section*{Model design}

The proposed model is based on the Visual Geometry Group Network (VGGNet, developed at the University of Oxford), a high-performing CNN model that is notable for its high depth and use of additional layers and small kernel size instead of fewer layers with a larger kernel size \cite{Simonyan2015}.
This helps to reduce the number of parameters of the network as well, especially important for small datasets which most models with many parameters struggle to converge on \cite{Kotsiantis2006}.

Specifically, the model is based on variation B of VGGNet (VGGNet B), with max pooling layers separated by groups of two convolutional layers with increasing filter count. Most notably, the filter count in each group of convolutional layers is smaller than for VGGNet B, ranging from 32 to 128 filters instead of 64 to 512. Additionally, the $2 \times 2$ max pooling layers are replaced with $2 \times 2$ average pooling layers. Max pooling is employed in most CNN models to increase contrast and preserve the most important information of an image while decreasing dimensionality of the input; however, contrast for the images is already quite high, with important features throughout the sample \cite{Nagi2011}. Average pooling retains more information from layer to layer because it incorporates all source pixels in the output compared to max pooling rejecting all but one of the pixels. When adaping VGGNet B to train on accelerometer data, this most significantly improved performance on the NIOSH lifting dataset. Table~\ref{bestmodel} contains a detailed description of the layers.

\begin{table}[!ht]
\centering
\caption{{\bf Detailed specification of the layers involved in the proposed model.} All 2D convolution layers contain a ReLU activation layer.}
\label{bestmodel}
\begin{tabular}{ll}
\multicolumn{2}{c}{\textbf{Model configuration }}  \\ 
\hline
\textbf{Layer type } & \textbf{Parameters }        \\ 
\hline
\multicolumn{2}{c}{Input (95x95x3 matrix)}         \\ 
\hline
2D convolution       & 32 filters, 3x3 kernel      \\ 
\hline
Average pooling      & 2x2 cell                    \\ 
\hline
Dropout              & 25\%                        \\ 
\hline
2D convolution       & 64 filters, 3x3 kernel      \\ 
\hline
2D convolution       & 64 filters, 3x3 kernel      \\ 
\hline
Average pooling      & 2x2 cell                    \\ 
\hline
Dropout              & 25\%                        \\ 
\hline
2D convolution       & 128 filters, 3x3 kernel     \\ 
\hline
2D convolution       & 128 filters, 3x3 kernel     \\ 
\hline
Average pooling      & 2x2 cell                    \\ 
\hline
Dropout              & 25\%                        \\ 
\hline
\multicolumn{2}{c}{Flatten}                        \\ 
\hline
Fully connected      & 1024 units                  \\ 
\hline
\multicolumn{2}{c}{Batch normalization}            \\ 
\hline
Dropout              & 25\%                        \\ 
\hline
\multicolumn{2}{c}{Softmax output}                 \\
\hline
\end{tabular}
\end{table}

In addition to the proposed model, a separate model (CNN+LSTM) was developed as an alternative approach, utilizing a network of 1-dimensional convolutional layers and long short-term memory (LSTM) layers, based on DeepConvLSTM by Ordóñez et al \cite{Ordonez2016}. This approach produces a far less complex network and treats the dataset as a time series instead of a 3D matrix thanks to the LSTM layers. LSTM layers utilize the current state of the network with their memory units, building direct relationships between the currently analyzed data and previously analyzed data. This makes it especially effective on temporally organized data, such as accelerometer and gyroscope signals \cite{Sak2014}.

The CNN+LSTM model also uses a slightly data format. Instead of the 97x97x3 matrix, the 12x750x3 matrix is utilized. Features are extracted by sliding a 12x1x3 window (essentially a column of the matrix) over the trial, reducing the dimensionality of the data from 3D to 2D and making it compatible with the LSTM layers. See \cite{Ordonez2016} for additional details on the structure of LSTM and 1D convolutional layers.

Three other models were developed strictly for comparison: a simpler CNN model with lower depth, a max pooling VGGNet B variation, and a multi-layer perceptron (MLP) model. The simpler CNN model does not utilize any form of pooling or a dense layer, consisting only of convolutional layers and a softmax output layer. The max pooling VGGNet B model is identical to the proposed model in all ways save the usage of max pooling layers instead of average pooling.

Models were all trained in the same manner, utilizing ADAM \cite{Kingma2015}
as a gradient descent estimator, categorical cross-entropy as a loss function
and, in place of a specific number of epochs, utilizing early stopping to halt training. The early stopping module monitored loss with a min delta of 0 and patience of 10 epochs. After loss failed to decrease, the best-performing (according to loss, not accuracy) weights of the last 10 epochs were selected for testing.

To develop the models, Keras v2.2.4 was used as an interface to TensorFlow v1.15.0. Data manipulation and standardization was performed with Scikit-learn v0.21.3.

\subsection*{Hyperparameter tuning}

A difficult portion when developing machine learning models is tuning the various hyperparameters, which are parameters for the model that are statically set before training begins. These parameters can affect training and testing results as significantly as alterations to the model structure and so deserve their own discussion.

In developing the model, three hyperparameters were focused on due to their great effect on accuracy: L2 regularization importance ($\lambda$), dropout percentage, and learning rate ($\alpha$). Searches were not performed for global minima due to the extreme effort involved and were tested with both single-value variation and multiple-value variation to evaluate any effects dependent on multiple hyperparameters.

$\lambda$ configures the importance of the L2 regularizer, which is applied solely to the final softmax dense layer in both the activity and output portions. Regularization in general is the practice of adding a loss function to the \textit{complexity} of the layer and incentivizing training of a sparser and less complex model. This helps to reduce the prevalence of overfitting to the training data \cite{Cortes2012}. Importance is the weight given to this regularizer over the basic loss function used in training; a higher value signifies higher importance. Most values for $\lambda$ are between 0.1 and $10^{-10}.$

Dropout percentage determines the portion of the weights at each layer that, for each epoch, are temporarily removed from the model and inserted after training the epoch. Dropout helps to further reduce overfitting by forcing the network to learn the same features multiple times as the portions of the model that previously trained are randomly removed \cite{Hinton2012}. Increased dropout can further prevent overfitting but also introduce instability when training, justifying tuning this parameter.

$\alpha$ configures the degree to which the optimizer tweaks the weights at each layer to descend the gradient. A higher value results in typically faster descent but increased instability. Overshooting the target, increasing the loss, and having to backtrack is more common. Lower $\alpha$ can help to prevent this, but typically requires a far longer training time and may lead to the optimizer converging poorly.

When comparing these models performance, a single value was desired to prevent subjectivity. However, traditional measures like F-measure and accuracy can fail to sufficiently capture performance, especially for multiclass and imbalanced datasets as present here. To correct for this and utilize a more descriptive statistic, R\textsubscript{K} correlation was utilized \cite{Gorodkin2004}. R\textsubscript{K} is a generalization of Pearson and Matthews correlation for multiclass problems. It has been shown to be superior to Cohen's kappa and standard metrics like f-measure (the latter specifically due to its inclusion of true negatives) as a single-metric comparison tool \cite{Delgado2019}.

Eq~\ref{eq:rk} displays a discretized version of R\textsubscript{K} that allows for its usage on a single confusion matrix \cite{Gorodkin2004}. $C_{lk}$ represents elements of the matrix.

\begin{eqnarray}
\label{eq:rk}
    R_{K} = 
        \frac{
            \sum \limits_{klm} C_{kk} C_{lm} - C_{kl} C_{mk}
        }{
            \sqrt{
                \sum\limits_{k} 
                \sum\limits_{l} C_{kl}
                \sum\limits_{\substack{l' \\ k' \neq k}} C_{k'l'}
            }
            \sqrt{
                \sum\limits_{k} 
                \sum\limits_{l} C_{lk}
                \sum\limits_{\substack{l' \\ k' \neq k}} C_{l'k'}
            }
        }
\end{eqnarray}

\section*{Results}

720 total trials were used for training and testing. Table~\ref{classcounts} lists the number of trials for each class label.

\begin{table}[!ht]
\centering
\caption{
{\bf Number of trials for each class of lift.}}
\begin{tabular}{ll} 

\textbf{ Class } & \textbf{ Count }  \\ 
\hline
Low Risk         & 120               \\ 
\hline
Medium Risk      & 240               \\ 
\hline
High Risk        & 360               \\
\hline
\end{tabular}
\label{classcounts}
\end{table}

\subsection*{Hyperparameter results}

\begin{figure}[!ht]
    \ifshowfigs
    \includegraphics{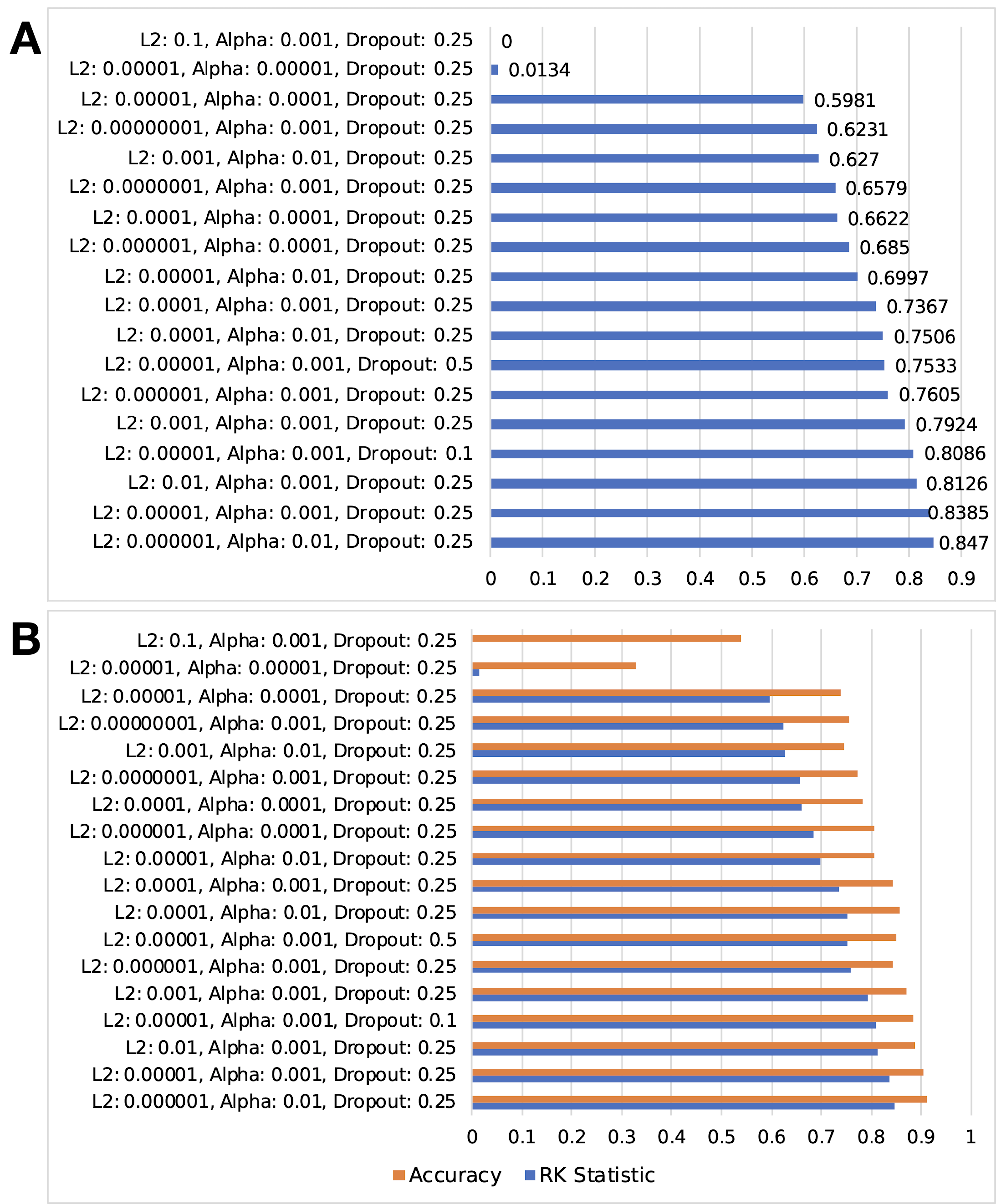}
    \fi
    \caption{{\bf Comparison of performance for various hyperparameters set on the proposed model.} A: comparison of R\textsubscript{K} statistics. B: R\textsubscript{K} statistic compared with accuracy. Accuracy ranges from 0 to 1 while R\textsubscript{K} ranges from -1 to 1. Values from -1 to 0 are not shown due to no results in that range.}
    \label{fig:rk} \label{fig:rkAcc}
\end{figure}

Fig~\ref{fig:rk} displays the results of training the proposed models with different values for the stated hyperparameters. Fig~\ref{fig:rkAcc} displays how R\textsubscript{K} can report different results due to its consideration of true negatives compared to accuracy, making it a more well-rounded single metric for comparing various models.

\begin{figure}[!ht]
    \ifshowfigs
    \includegraphics[width=\textwidth]{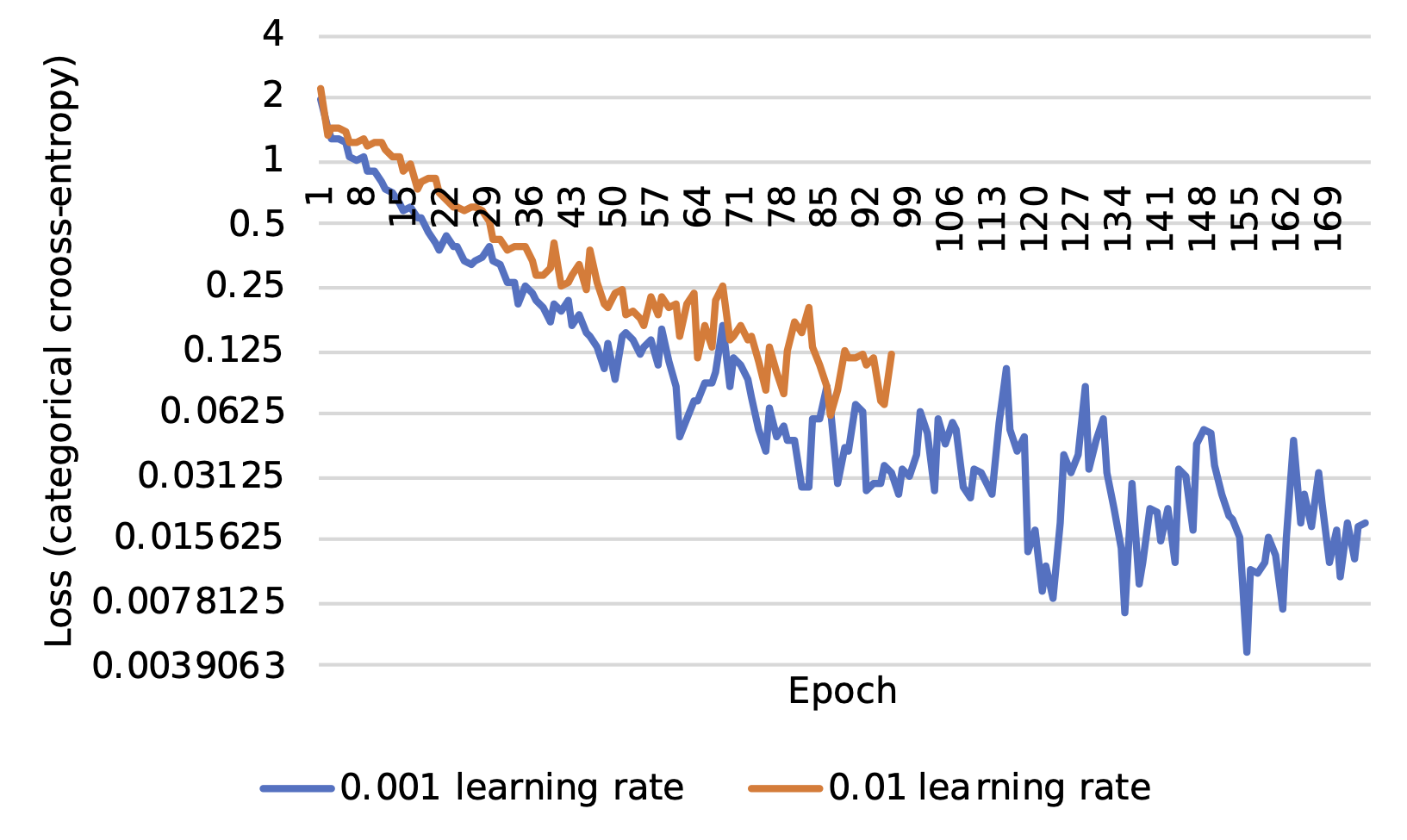}
    \fi
    \caption{{\bf Loss gradient descent for $\alpha$ of 0.01 and 0.001.} Each point represents the training categorical cross-entropy at the completion of each epoch.}
    \label{fig:alpha}
\end{figure}

However, when selecting the resulting hyperparameters, a qualitative selection of the second-best performing parameters was made. This is due to the behavior of the optimizer at such a high learning rate; it behaved in a more unstable pattern and terminated at a relatively high categorical cross-entropy compared to a slightly lower $\alpha$. Fig~\ref{fig:alpha} displays the behavior involved. Therefore, the parameters utilized for the proposed model and variant with max pooling are $\lambda = 10^{-5}, \alpha = 10^{-3}, \text{dropout percentage} = 25\%$.

\subsection*{Classification results}

Although various model structures were used, all relied on the same basic layer unit: the 2D convolutional layer. Models varied in complexity and depth, with some containing LSTM layers in an attempt to formally capture time-based data.

Table~\ref{modelcomparison} compares precision, recall, and f-measure for the proposed model, a typical simplified CNN model, and a fully-connected multilayer perceptron network.

\begin{table}[!ht]
\begin{adjustwidth}{-2.25in}{0in} 
\centering
\caption{Summary of classification results for proposed model and alternatives.}
\label{modelcomparison}
\begin{tabular}{|l|l|l|l|l|l|} 
\hline
 \textbf{Risk Level}        & \textbf{Proposed Model}  & \textbf{CNN+LSTM }  & \textbf{VGGNet B (Max Pool) }  & \textbf{CNN }  & \textbf{MLP }   \\ 
\hline
 \textbf{Low F-measure}     & 0.776                    & 0.8                 & 0.746                          & 0.444          & 0.367           \\ 
\hline
 \textbf{Medium F-measure}  & 0.862                    & 0.783               & 0.692                          & 0.635          & 0.500           \\ 
\hline
 \textbf{High F-measure}    & 0.964                    & 0.923               & 0.815                          & 0.765          & 0.667           \\ 
\hline
 \textbf{Accuracy}          & \textbf{0.906}           & 0.861               & 0.766                          & 0.667          & 0.567           \\ 
\hline
 \textbf{RK}                & \textbf{0.839}           & 0.77                & 0.65                           & 0.53           & 0.28            \\
\hline
\end{tabular}
\end{adjustwidth}
\end{table}

\begin{figure}[!ht]
    \ifshowfigs
    \includegraphics[width=\textwidth]{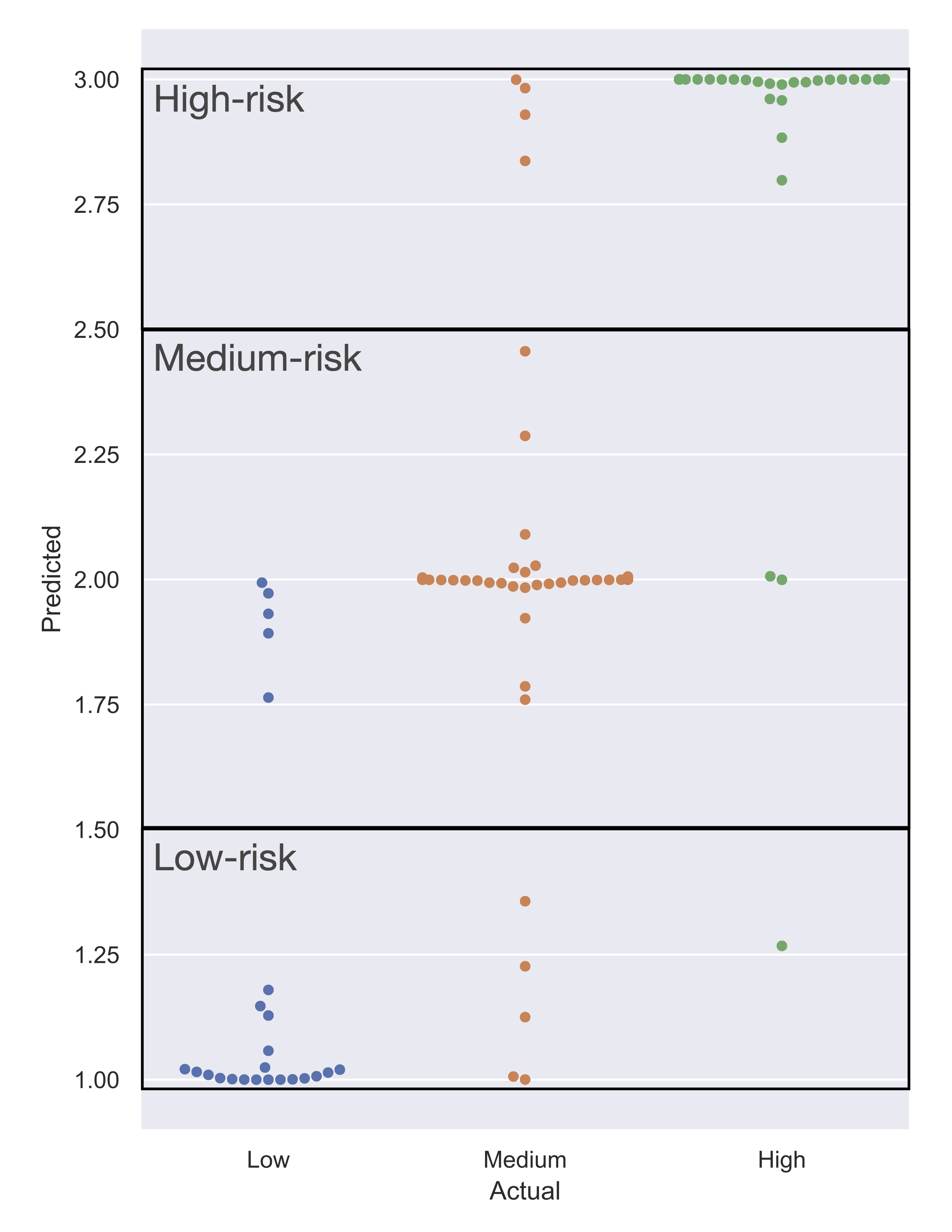}
    \fi
    \caption{{\bf Swarm plot of the testing results by the proposed model.} The x-axis represents the true labeling and the y-axis the model output. The y-axis is divided into three zones that define the resulting class for each value, labeled in the top-left of each box.}
    \label{fig:swarmplot}
\end{figure}

\begin{figure}[!ht]
    \ifshowfigs
    \includegraphics[width=\textwidth]{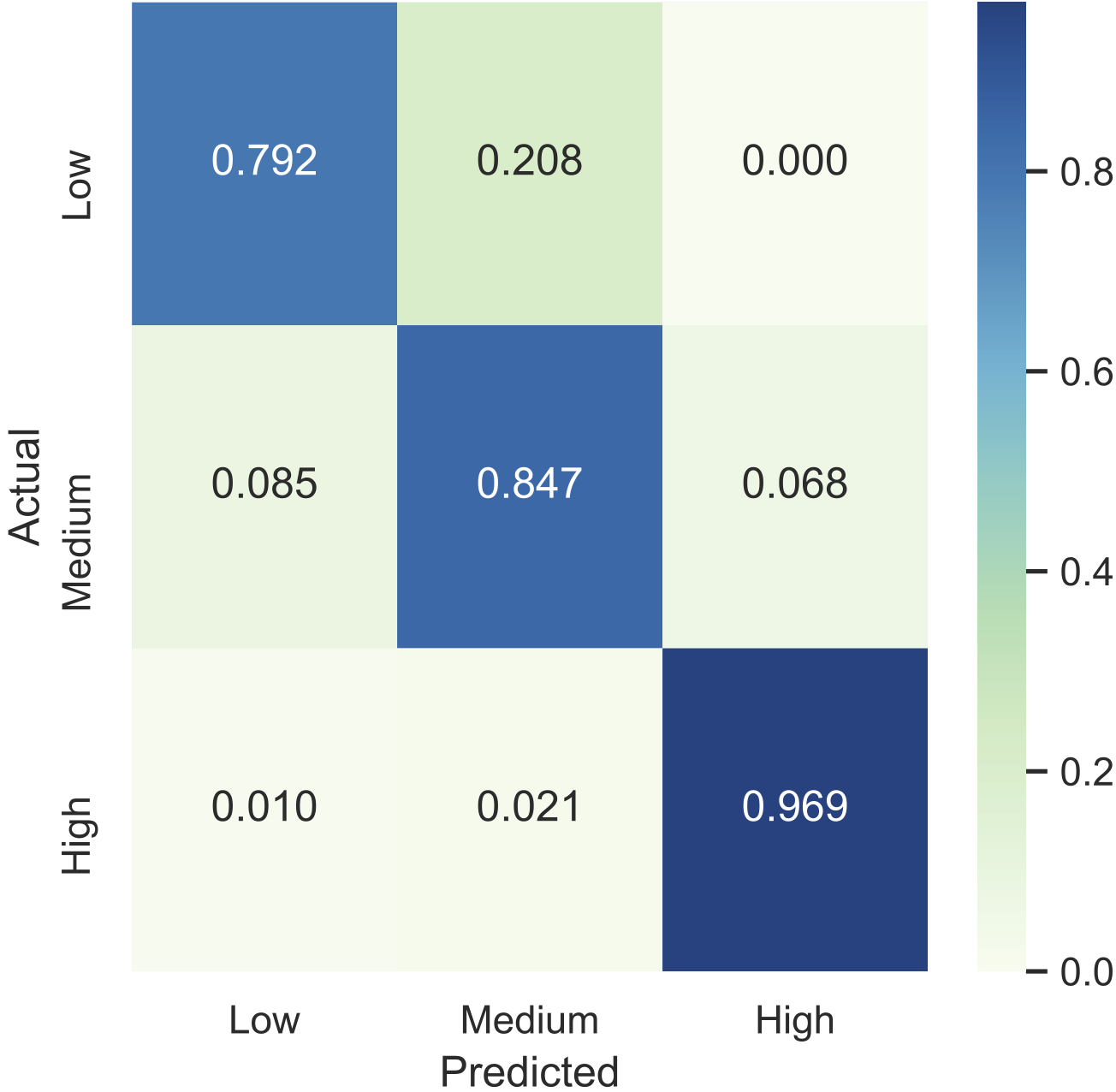}
    \fi
    \caption{{\bf Heatmap plot of the testing results by the proposed model.} Each row has been normalized so that each class has the same color scale.}
    \label{fig:heatmap}
\end{figure}

The modified VGGNet B model performed the best out of all models on medium and high-risk lifts and only slightly underperformed the max pooling variant in low-risk lifts. It also significantly outperformed the simplified CNN and multilayer perceptron in f-measure for all classes and in overall accuracy. Fig~\ref{fig:swarmplot} displays the distribution of predictions for the proposed model on the testing data. The results for the proposed model are displayed in more detail in Fig~\ref{fig:heatmap}.

\section*{Discussion}

Small datasets are notoriously difficult to train on for machine learning models and that is borne out here, with typical CNN and MLP models barely outperforming randomness for some guesses and delivering no more than a 2/3 accuracy. The proposed VGGNet B variant significantly outperformed the other two models and max pooling variation, both in R\textsubscript{K} (0.862) and accuracy (90.6\%). Notably, it also displayed the best f-measure for every single class as well, although it improved the least in low-risk classification.

As R\textsubscript{K} is typically only considered satisfactory above an 0.7 correlation, the proposed model and CNN+LSTM models were also the only methods that displayed acceptable performance.

Notably, the data trained on (see Table~\ref{classcounts}) is an example of an imbalanced dataset, with significantly varying numbers of samples in each risk level. This typically poses a challenge for machine learning models that automatically become biased toward the larger class \cite{Kotsiantis2006}. This is obvious from examining the f-measures for all models, which increase with the number of trials for the class.

While low-risk lifts were classified more poorly than other classes, the proposed model improves upon all alternatives and helps to avoid issues such as over and under-sampling, which can lead to worse performance when testing \cite{Kotsiantis2006}.

\subsection*{Average pooling}

Average pooling, other than parameter changes, is the major departure from VGGNet B for the model. It resulted in a 5.3 basis point increase in accuracy with major benefits to both medium and high-risk accuracy. It is theorized that, in this case, average pooling outperforms max pooling because it passes more information to the next layer by using all 4 cells in the pool instead of selecting the highest value. This could increase generalizability of the model to testing data. However, max pooling is typically the selected model for successful CNN-based classifiers \cite{Simonyan2015}.

These models typically are trained on largely unprocessed image data that may have low contrast and significant redundancy in a given region, reducing the information loss of max pooling. Max pooling picking a blue pixel from the sky in an image is still representative. Average pooling is similar to various image downsampling methods, albeit simplified in its attempts to preserve information. As accelerometer data, especially at low frequencies (the dataset was recorded at 25 Hz), is vulnerable to loss of information, preserving this information may have resulted in the increase in accuracy. Use of max pooling would simulate a very rough downsampling of the data, which could clip many of the important fluctuations in rotation and acceleration from the sensor data. Average pooling smoothes out this downsampling and, while it still eliminates information, performs it less severely than max pooling.

\subsection*{Saliency mapping}

Saliency mapping is the process of determining the input features that the model recognizes as the most significant to the output class. First defined by Simonyan et. al. in \cite{Simonyan2014}, saliency is a multistep process. Given a final score matrix $S_{c}(I)$ for a class $c$ selected by the model, the linear score can be represented in Eq~\ref{eq:linear_score} as

\begin{eqnarray}
    S_{c}(I) = w_{c}^{T} I + b_{c}
    \label{eq:linear_score}
\end{eqnarray}

where $I$ is the pixels of the image, $w_{c}$ is the weights for the class, and $b_{c}$ is the bias of the class. As the model is non-linear in nature due to the activation functions and so could not be easily computed, the approximation is computed instead in Eq~\ref{eq:saliency-derivative}. 

\begin{eqnarray}
    S_{c}(I) \approx w^{T} I + b
    \label{eq:score-approx}
\end{eqnarray}

$w^{T}$ is the derivative of the score matrix $S_{c}$ to the image matrix $I$ at the point $I_{0}$ (the image itself), shown in Eq~\ref{eq:saliency-derivative} \cite{Simonyan2014}.

\begin{eqnarray}
    w = \frac{\delta S_{c}}{\delta I} \bigg\rvert _{I_{0}}
    \label{eq:saliency-derivative}
\end{eqnarray}

This approximation determines how much each pixel, if changed, would affect the class score and assigns a value to them.

Here, the primary purpose was to determine whether the model was truly recognizing portions of the input as the lift or simply tweaking the weights to fit on noise, which is a possibility. However, this is not the behavior displayed in the saliency plots obtained from the proposed model. 

\begin{figure}[!ht]
    \ifshowfigs
    \includegraphics[width=\textwidth]{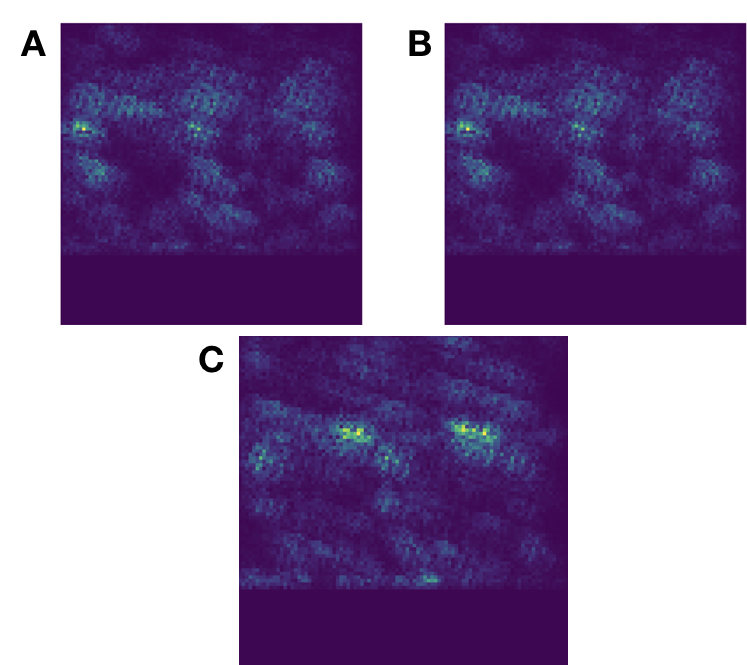}
    \fi
    \caption{{\bf Saliency plots for final softmax layer of network.} A: low-risk saliency. B: medium-risk saliency. C: high-risk saliency. Bright green/yellow represents the highest weighting; dark purple represents the lowest weighting.}
    \label{fig:saliency} \label{fig:saliency_low} \label{fig:saliency_med} \label{fig:saliency_hi}
\end{figure}

The low and medium saliency plots (Fig~\ref{fig:saliency_low} and Fig~\ref{fig:saliency_med}) are nearly identical but this is not an indicator that they cannot be differentiated. Instead, the model considers the same parts of the lifting motion as important for each class. For high-risk lifts, shown in Fig~\ref{fig:saliency_hi}, the model is drawing on two very specific regions, with very low weighting for the rest of the image. We theorize that it may have determined two specific points of the lift: when the object is lifted and when the lifter accelerates back to a neutral, upright position. These regions are also present on the other two saliency plots but are not as clearly delineated. This difference in quality is the clearest indicator that further improvements are possible beyond simply examining classification accuracy. However, this would most likely require alterations to the structure of the model and not simple hyperparameter tuning.

Producing these plots also provides a separate benefit: examining what sensors contribute most effectively to the results. While more sensor data can assist in classification, many of the sensors have difficult or impractical placement, such as those on the thigh and upper back. On the high-risk classification, the two hot areas center around the wrist and back sensors. Side, upper arm, and thigh sensors contributed to the classification but, as shown in Fig~\ref{fig:saliency}, are not as bright and so are candidates for possible removal in future research.

\begin{figure}[!ht]
\begin{adjustwidth}{-2.25in}{0in}
    \ifshowfigs
    \centering
    \includegraphics[width=7.5in]{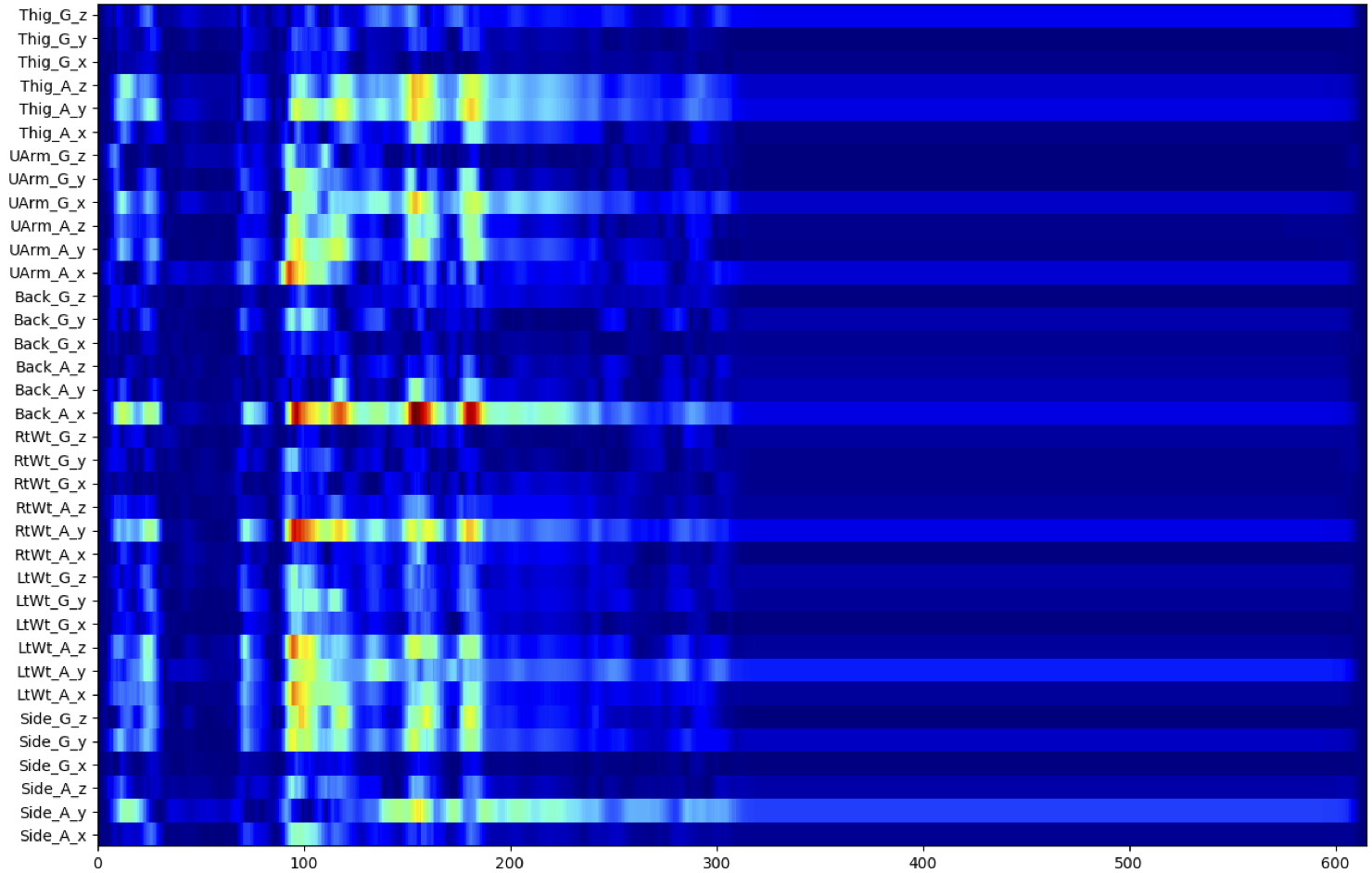}
    \fi
    \caption{{\bf Saliency plot for high-risk lift trials obtained from the CNN+LSTM model.} Scale ranges from deep blue as the lowest significance and deep red as the highest significance. The x-axis is frames of the input data and y-axis is the sensor data, where A/G is accelerometer or gyroscope and x, y, z are the dimensions for the sensor.}
    \label{fig:saliency_hi_lstm}
\end{adjustwidth}
\end{figure}

A significant advantage of the CNN+LSTM model is, due to the data input shape, saliency analysis is far clearer here. Fig~\ref{fig:saliency_hi_lstm} displays the saliency for high-risk lifts as well, with the various sensors labeled. Here, it is far clearer that the back and wrist sensors are the most significant. While all sensors contribute to the prediction (in at least one dimension), removing the thigh, upper arm, and side sensors may still leave enough information to sufficiently classify lifting risk level.

Additionally, we see two general peaks here as well, suggesting that lifting behavior may have an initial acceleration and final acceleration as general features. As both models appear to be examining the same region of the data, it is increasingly likely that the data is sufficiently separable and contains true features instead of simple noise.

\subsection*{Feature extraction}

Many other examples of accelerometer classification employ manual feature extraction with various measurements including means, variances, zero-crossing rates, and various other statistics \cite{Arif2015}.

This method only requires a signal filter and standardization before reshaping and feeding to the model. This could be performed very quickly and benefits from significant previous efforts in signal processing to allow streamed and real-time transformations. While many more features are utilized, handcrafted features typically require significant domain experience and tuning for high performance. The preprocessing proposed could easily be applied to many different datasets with minimal alterations.

\subsection*{Possible improvements and future research}

90.6\% accuracy, while a significant improvement over other models, still provides opportunity for further advancement. It is doubtful that hyperparameter tweaking could significantly increase accuracy, as Fig~\ref{fig:rkAcc} displays only small increases once 85\% accuracy was reached. Model alterations are most likely necessary to reach overall 95\% testing accuracy. However, high-risk classification, considered most important to ensure worker safety, is excellent, with 96.9\% accuracy.

One possible region of interest is configuring the number of convolutional layers between the average pooling layers. Deeper into the network, additional convolutional layers may provide high-level feature extraction. Nearer to the input layer, more layers typically increase recognition of granular details. However, as stated in the introduction, a high number of parameters for a small dataset such as this can result in the optimizer failing to converge. Therefore, simply adding layers may fail to significantly increase accuracy and require alterations to the general network structure as well.

The presence of a fully-connected layer at the end of the network is also a point of interest. This layer especially provides a significant number of parameters and so is a target for optimization. Altering the number of units and configuring a regularizer on that layer may assist, especially on improving low-risk lifts due to optimizing for a less complex model.

For real-world use, minimizing the number of sensors will significantly advance the practicality, reducing cost and eliminating the awkward placement of several sensors. Ideally, one sensor would provide sufficient data to classify lifting. However, the two wrist sensors and a side sensor are simple enough to attach that they may be an acceptable alternative. The strong emphasis, unfortunately, by both models on the back sensor may indicate that this sensor is necessary. Redundancies may nonetheless exist in the sensor data and may provide sufficient separability for a similar model.

Finally, based on the saliency maps generated, the window size for the lift could be significantly reduced with, in all likelihood, minimal reduction in accuracy. This would increase the speed of activity recognition in real-world use simply due to the reduced input size. A reduction in input size also decreases the number of parameters in the network by eliminating features, possibly allowing for further depth in the network. However, overzealous input size reduction could cut off significant parts of lifts and so needs to be performed carefully.

\section*{Conclusion}

Classifying accelerometer data is traditionally difficult due to the requirement of most machine learning models requiring large datasets. Therefore, much of the existing research focuses on typical activity and exercise classification that can draw on existing datasets or be compiled from many subjects. Specialized activities with lesser impact, then, have been neglected due to the difficulty involved in compiling enough data for traditional models.

The proposed model was able to quickly and accurately (90.6\% accuracy, 0.839 R\textsubscript{K}) classify a small accelerometer dataset provided by NIOSH with minimal feature extraction and significantly greater performance than other models tested. Specifically, the usage of a CNN that would normally classify images along with the alteration to use average pooling over max pooling provided the greatest benefit. Hyperparameter tuning was also shown to have significant effects on the performance, but in lower magnitudes. It is very likely that a similar model could be trained on other small and/or unbalanced datasets to make their classification feasible where other models have failed. The proposed technique provides excellent monitoring of risks involved in the various types of manual lifts in an industrial setup to provide timely interventions.

\section*{Acknowledgments}
This study was funded, in part, by the National Institute for Occupational Safety and Health (NIOSH) and the Research Experiences for Undergraduates Program of the National Science Foundation (REU NSF) REU supplement for NSF award no.: ECCS 1556294. Disclaimer: the findings and conclusions in this report are those of the authors and do not necessarily represent the official position of NIOSH, Centers for Disease Control and Prevention (NIOSH/CDC) or NSF.  Mention of any company or product does not constitute endorsement by NIOSH/CDC or NSF.

%
%
%

\end{document}